\documentclass[]{article}

\usepackage{arxiv}

\usepackage[utf8]{inputenc} 
\usepackage[T1]{fontenc}    

\usepackage{amsfonts}       
\usepackage{nicefrac}       
\usepackage{lipsum}		

\usepackage[square,sort,comma,numbers]{natbib}

\usepackage{doi}

\usepackage[onehalfspacing]{setspace}

\usepackage{booktabs}
\usepackage{tabularray}
\usepackage{tabularx}
\usepackage{siunitx} 
\usepackage{arydshln}

\usepackage{nicematrix}

\usepackage{graphicx}

\usepackage{hyperref}
\usepackage{url}
\usepackage{subfig}
\usepackage{caption}
\usepackage{subcaption}
\usepackage{float}
\usepackage{amsmath}
\usepackage{hyperref}

\usepackage{authblk}

\author[]{Zahra Sadeghi
\thanks{Correspondence: zahras@dal.ca} }
\author[] {Stan Matwin}
\affil{Faculty of Computer Science, Dalhousie University, Halifax, Canada}
 
\begin{document}
\onecolumn

\title{A Review of Global Sensitivity Analysis Methods and a comparative case study on Digit Classification}

\date{\vspace{-5ex}}

\maketitle

\begin{abstract}
Global sensitivity analysis (GSA) aims to detect influential input factors that lead a model to arrive at a certain decision and is a significant approach for mitigating the computational burden of processing high dimensional data. In this paper, we provide a comprehensive review and a comparison on global sensitivity analysis methods. Additionally, we propose a methodology for evaluating the efficacy of these methods by conducting a case study on MNIST digit dataset. Our study goes through the underlying mechanism of widely used GSA methods and highlights their efficacy through a comprehensive methodology. 
\end{abstract}

\section*{Introduction}
In the era of deep learning and the rapid advancement of powerful Artificial Intelligence (AI) models, consisting of numerous layers and millions of parameters, the demand for understanding the decision-making process of black box models is on the rise. Explainable AI is a growing trend that seeks to uncover the inner workings of AI systems through computational analysis, shedding light on the decision-making process and has been applied across a variety of data types such as video \cite{aakur2018inherently}, text \cite{sarvmaili2022modularized}, AIS \cite{veerappa2022validation} and causal \cite{taylor2024causal} and genomic data \cite{van2022comparison}, and applications such as art \cite{bryan2023explainable}, medicine \cite{sadeghi2024review}, finance \cite{weber2024applications} and education \cite{fiok2022explainable}.
Explainability methods can be broadly divided into model agnostic or model free and model specific approaches. Model-agnostic methods can be applied to any trained machine learning model regardless of the learning mechanism and model architecture. Rule based methods \cite{vilone2020comparative} and sensitivity analysis are two common approaches from this category. Statistical analysis and sensitivity analysis are two common model agnostic approaches. 
Statistical methods such as Spearman rank correlation coefficient (SPEA), Standard regression coefficient (SRC), Partial Correlation Coefficient (PCC), Pearson correlation coefficient \cite{gan2014comprehensive} can be referred to as a few approaches which are employed for finding the association between input parameters and model output. Sensitivity analysis methods refer to a broad approach for examining relationships between a system's inputs and outputs. According to Saltelli et al., sensitivity analysis investigates the process of how the uncertainty in a model's output can be attributed to different sources of uncertainty within the model's inputs \cite{saltelli2004sensitivity}.

In contrast to model-agnostic techniques, model-specific methods are tailored for and suitable only for particular machine learning models, such as those developed to explain the inner workings of deep neural networks. Consequently, a substantial body of research is dedicated to developing explanations of the predictions made by deep learning models through network dissection \cite{zhou2018interpreting}, gradient optimization and visualization \cite{selvaraju2017grad} \cite{shrikumar2017learning}, saliency maps and visualization \cite{simonyan2013deep} \cite{sadeghi2019information}, loss landscapes \cite{li2018visualizing} \cite{barrett2023evolutionary}, and semantic analysis \cite{pavlick2022semantic}, \cite{jingjing2022semantic},\cite{sadeghi2016deep}.

From another perspective, explainability methods can take a global or local processing approach. Global explainability methods focus on explaining the overall behavior of a machine learning model by varying the entire range of input factors and examining the joint effect and interaction between them. In this approach, all input parameter are allowed to change simultaneously across all the possible range of values. In contrast, local explianibilty methods study the effect of a parameter by exploring its local vicinity while holding all other parameters fixed at their baseline values \cite{li2023comparison} \cite{qin2023comparative}. Local interpretability methods typically make the underlying assumption that the machine learning model exhibits nonlinear relationships and independence between the input parameters. Consequently, if the input factors exhibit significant interactions, local interpretability methods may produce misleading or inaccurate results, as they rely on the assumption of parameter independence \cite{reed2022addressing}. The aforementioned statistical method can be considered as local approaches. 
Shapley values \cite{chen2021explaining} and LIME (Local Interpretable Model-Agnostic Explanations) \cite{ribeiro2016should} are two of the most widely-used local interpretability approaches. In the context of sensitivity analysis, local SA methods examine the input-output relationship at specific, individual points within the model's parameter space \cite{saltelli2004sensitivity} \cite{saltelli2002sensitivity}. 
Global interpretability methods can be applied without the need to make assumptions about the relationships between the input parameters. Another advantage of global interpretability methods is that the trained models can be analyzed as black box models with relatively minimal coding effort required. However, there is no universal consensus on how to measure the impact of parameters using global interpretability methods. Different global techniques are grounded in varying mathematical foundations. As a result, they can produce divergent rankings or measures of parameter importance when applied to the same machine learning model. Therefore, global interpretability methods can be applied for analyzing the effect of parameters on the model overall output, whereas local techniques are applicable for studying this effect for individual cases. 
Example of global approaches include Partial Dependence Plot (PDP) \cite{molnar2022interpretable} and  (Individual Conditional Expectation) ICE \cite{molnar2020interpretable}. In this paper, we explore global sensitivity analysis (GSA) methods in depth and compare their effectiveness through a case study. This paper is structured as follows: We first review the global sensitivity analysis methods, next, we propose a methodology for comparing the performance of the discussed GSA methods. We present the results and conclude the paper in the last section. 

\section {Global Sensitivity Analysis methods}\label{sec:sec_method}
The general paradigm of global sensitivity analysis (GSA) methods consists of two phases of sampling and analysis. Initially, a set of samples are generated from a variable or factor $X_i$. Next the output vector $Y$ is produced using a model $f$ for all the variables:
\begin{align}
Y = f(X_1,...,X_p)
\end{align}
Finally, the impact of each variable is analyzed and assessed. 
GSA methods can be roughly categorized into four different groups, i.e, variance based methods, derivative based methods, density based methods and feature additive methods.
\subsection{Variance based methods}
Variance based methods are based on the assumption that variance is sufficient to describe the output uncertainty, an assumption made by Saltelli et al. \cite{sobol2001global} \cite{saltelli2002making} \cite{saltelli2010variance}. The rationale behind this approach is based on the variance of the expected output value conditioned on parameters. A higher variance suggests a lower level of importance for a parameter.
\subsubsection{Sobol}
Sobol \lq s method relies on decomposition of the model output variance under the assumption that inputs are independent and uncorrelated. Equation (1) to (3) illustrate the variance decomposition of variable $Y$ in Sobol\lq s approach.  
\begin{align}
V(Y) = \sum_{i=1}^pV_i + \sum_{1 \leq i < j \leq p}^{}V_{ij}+. ..+V_{1,...,p}
\end{align}
\begin{align}
V_i = V(E(Y|X_i))
\end{align}
\begin{align}
V_{ij} = V(E(Y|X_i, X_j))-V_i - V_j
\end{align}
Sobol sensitivity analysis assesses the impact of each input parameter, both in isolation and in conjunction with other parameters, resulting in the derivation of first-order and higher-order sensitivity indices. First‐order (S1), second‐order (S2), total‐order (ST) and higher‐order sensitivity indices are calculated to accurately reflect the influence of the individual input, and the interaction between them.
Equations (2) and (3) measure the first-order and second-order variations, while equations (4) and (5) show the first-order and second-order sensitivity indices, which indicate the fraction of variance of $Y$ caused by factor $X_i$ and the interaction between factors $X_i$ and $X_j$ respectively. 

\begin{align}
S_{i}=\frac{V_{i}}{V(Y)}
\end{align}
\begin{align}
S_{ij}=\frac{V_{ij}}{V(Y)}
\end{align}
Additionally, total-order sensitivity index is explained by equation (6) and is the summation over all the the variances induced by factor $X_i$ (denoted by $\#i$).
\begin{align}
ST_{i}=\sum_{k\in \# i}{S_k}
\end{align}

\subsubsection{FAST}
The Fourier amplitude sensitivity test (FAST) is based on periodic search sampling using a period search function and applies a decomposition of variance based on Fourier Transform. FAST describes the decomposition of variance using transfer functions as indicated by Equation (7): 
\begin{align}
X_i(s_j) = G_i(\sin(\omega_is_j)), i = 1, 2, ..., k, j = 1, 2, ..., n,  s\in(-\pi, \pi)
\end{align}
Where $\omega_i$ are integer frequencies.
The model output can be obtained by:
\begin{align}
Y = f(s) = f(X_1(s),...,f(X_k(s))
\end{align}
The variance of Y is then defined according to Equation (9) and approximated by Equation (10) by incorporating Parseval's theorem:
\begin{align}
V(Y) = \frac {1}{2 \pi} \int_{-\pi}^{\pi}f^2(s)d(s)-[E(Y)]^2
\end{align}
\begin{align}
V(Y)\approx2\sum_{p=1}^\infty(A_p^2+B_p^2)
\end{align}
The first order sensitivity indexes of FAST are then computed by employing Equations (11).
\begin{align}
S_i=\frac{V_i}{V(Y)}\approx\frac{\sum_{q=1}^M(A_{q \omega_i}^2+B_{q \omega_i}^2)}{\sum_{i=1}^n\sum_{q=1}^M(A_{q \omega_i}^2+B_{q \omega_i}^2)}
\end{align}
The total order sensitivity index or total effect of FAST can be obtained by considering all the first order effects as well as all the higher order effects that exclude index $i$.
\begin{align}
    ST_i = S_i + S_{i, \sim i}
\end{align}
FAST achieves a better estimate in terms of robustness and speed of convergences than Sobol and can be applied to  nonlinear and non-monotonic models. \cite{chan1997sensitivity} \cite{xu2008general}. 

\subsubsection{RBD and FAST\_RBD}
When the number of inputs increases, FAST encounters numerous sources of error and it results to poor estimation considering the computational cost of deriving all the higher order terms. The RBD and hybrid FAST\_RBD (aka HFR) methods developed to overcome this computational burden of FAST.

In contrary to FAST which explores the space by using different frequencies for each parameter (i.e., $\omega_i$), RBD takes a single frequency (i.e, $\omega$) for all the parameters which can be determined randomly and can be set to 1 for the sake of simplicity \cite{tarantola2006random}. 
This reduces the computational complexity of the algorithm, however, the search curve cannot cover the whole space completely. In order to avoid this problem, random permutation of the coordinate of sample points (i.e,. design points) are employed to boost the stochastic nature of the algorithm \cite{tissot2012bias}.
\begin{align}
X_i(s_ij) = G_i(\sin(\omega s_{i_j})), i = 1, 2, ..., k, j = 1, 2, .. ,n , S\in(-\pi, \pi)
\end{align}

In hybrid FAST, the $k$ parameters are grouped into partitions of equal size while a particular frequency is assigned to each partitions. Hence, HFR strikes a balance between the accuracy of FAST and the computational efficiency of RBD \cite{tarantola2006random}. 

\subsection{Derivative based methods}
This approach is based on determining the sensitivity indexes by computing the first order partial derivative of the model output with respect to the input variables ($\left | \frac{\partial  f}{\partial  x_i} \right |$). A higher derivative value suggests a greater level of sensitivity. Two significant methods falling under this category are Morris and DGSM.

\subsubsection{Morris}
The basic idea of Morris method is built upon calculation of elementary effects (EE) for each input factor by dividing the range of each factor into $p$ levels and considering $\Delta$ as a predetermined multiple of $\frac{1}{(p-1)}$, where $X_i + \Delta \leq 1$ for exploring the grid space: 

\begin{align} 
EE_{i}=\frac{[F(X_1, ..., X_{i-1}, X_i + \Delta, X_{i+1}, ..., X_k)-F(X_1, ..., X_{i-1}, X_i, X_{i+1}, ..., X_k)]}{\Delta_i}
\end{align}

After sampling $X_i$ and computing $EE_i$ $r$ times, the average and standard deviation of elementary effects are obtained which are known as the $\mu$ and $\sigma_i$ measurements:
\begin{align}
\mu_i = \frac{1}{r} \sum_{i=1}^rEE_{i}
\end{align}

\begin{align}
\sigma_i = \sqrt{\frac{1}{r}\sum_{i=1}^r\left(EE_{i} - \frac{1}{r} \sum_{i=1}^r (EE_{i})\right)^{2}}
\end{align}
Campolongo et al., proposed a revised version of $\mu$ called $\mu^*$ by considering the absolute value of $EE_i$ in order to mitigate the issue of cancellation of opposite signs in non-monotonic models \cite{campolongo2007effective}.
The higher values of $\mu_i^*$ indicate a greater influence of $X_i$  on the output, while higher values of $\sigma_i$ suggest increased interaction between $X_i$ and the other variables or a non-linear effect \cite{morris1991factorial}.
\subsubsection{DGSM} 
Morris method of importance measurement is based on approximation of $\int_{H^n}^{}\left(\frac{\partial f}{\partial x_i}\right)dx$, where $H^n$ is the sample space. The derivative based global sensitivity index is a generalization of Morris and is measured as:
\begin{align}
v_i=\int_{H^n}^{}\left(\frac{\partial f}{\partial x_i}\right)^2dx 
\end{align}
A low value of $v_i$ signifies non-important factors of low influence \cite{sobol2010derivative}. It is also shown that there is a link between DGSM values and Sobol's total index \cite{sobol2001global}.
\subsection{Distribution based methods}
The distribution based (or moment-independent) approach investigates the entire distribution of a model measured by its PDF and finds the sensitivity of the model based on the variation to its density function without relying on any moments of the output.
\subsubsection{DELTA}
Delta is a density based sensitivity method.
It assumes that all input variables are independent. However, this method can be used in the presence of correlation among the variables.
The idea behind this method is to find the shift $s$ between the unconditional density (or cumulative distribution) of $Y$ (i.e., $f_Y(y)$) and the conditional density  of $Y$ given that one variable $X_i$ is fixed with a constant value $x$ (i.e., $f_{Y|X_i=x}(y)$). The Delta sensitivity index consider the expected value of $s$:
\begin{align}
E_{X_i}[s(X_i)]=\int_{x}^{} f_{X_i}(x_i)\left[\int_{y}^{}|f_Y(y)-f_{Y|X_i}(y)|dy\right]dx_i
\end{align}
Where $f_{X_i}(x_i)$ is the marginal density of $x_i$. The sensitivity index of factor $X_i$ is then measured by $\delta$:
\begin{align}
\delta_i = \frac{1}{2}E_{X_i}[s(X_i)]
\end{align}
This Delta sensitivity index remains unaffected by monotonic transformations and is normalized ($0 \leq \delta_i \leq 1$) \cite{borgonovo2007new}.  

\section*{Method}
In this section, we propose a method for making a comparison between the performance of discussed sensitive analysis methods in the previous section. Our study leverages the machine learning simulations for Sensitivity analysis which focuses on feature selection and classification accuracy as a basis for evaluation.To this end, we first train a simple two-layer Convolutional Neural Network (CNN) on MNIST data, where we employ a 2D convolutional operator followed by ReLU and Maxpooling. We trained this network on MNIST training data and achieved 99\% classification accuracy. Then we apply each of the SA methods using this trained CNN on the MNIST test data to obtain the most important features from each of the images. We seek to understand the contribution of digits' visual patterns for correctly predicting their category by measuring the effect of the important features as determined by each of the SA methods.

 To achieve this objective, we propose a methodology to evaluate the performance of important features on classification and clustering tasks. For this purpose, the important features of each image are first sorted and categorized into seven blocks. The initial block comprises all the 784 pixels of an image from MNIST dataset, while subsequent blocks gradually decrease by 100 pixels until reaching the final block, which contains 84 most important pixels. We study the efficacy of both the important and non-important pixels.
 
\begin{figure}[ht]
     \centering
     \includegraphics[scale=0.45]{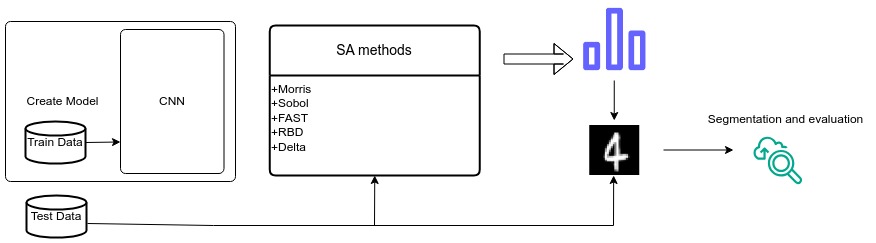}
     \caption{Methodology for comparing SA methods}
     \label{fig:diagram}
 \end{figure}

 We subsequently assess the classification and clustering performance of the test data solely based on the pixel information on each of the chunks. Through this approach, we can measure the impact of both the influential and non-influential pixels on digit detection task. \autoref{fig:diagram} illustrates the methodology procedure. For implementation we applied Sensitivity Analysis Library in Python (SALIB \footnote{https://salib.readthedocs.io/en/latest/index.html}). 
 
\section*{Results and conclusions}
We evaluate the effect of the important features determined by each of the SA algorithms discussed in \autoref{sec:sec_method} on the accuracy of digit prediction and compare the influence of important and not-important features as reflected by each of the SA methods. The hyperparameters which are selected for each method is listed in Table \autoref{tab:table1} which are determined by grid search. The number of samples is listed in Table \autoref{tab:table1} and is determined using a grid search, which strikes a balance between optimizing performance and minimizing the number of samples required.

\begin{figure}[htbp]
 \centering
  \subfloat[]  {\includegraphics[width=0.45\textwidth]{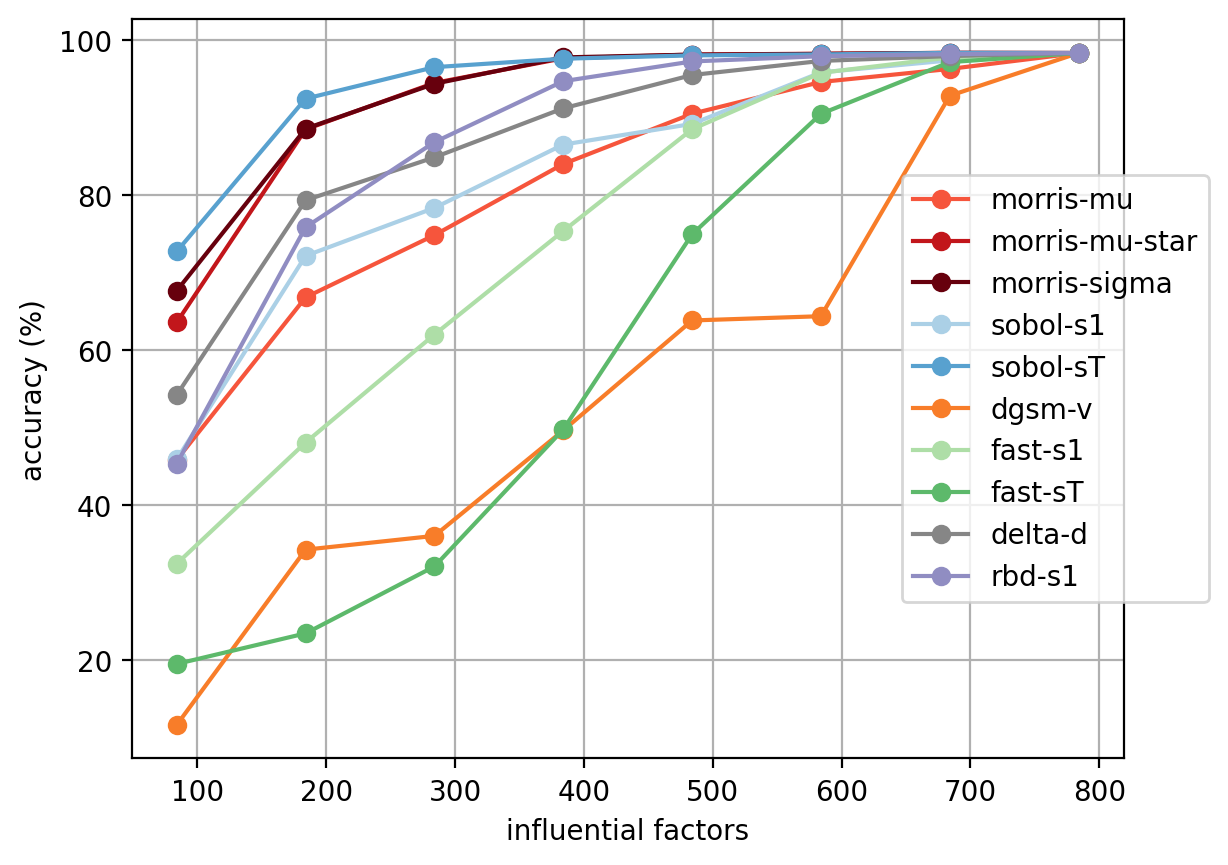}}\label{fig:fig2_a}
  \subfloat[]{\includegraphics[width=0.45\textwidth]{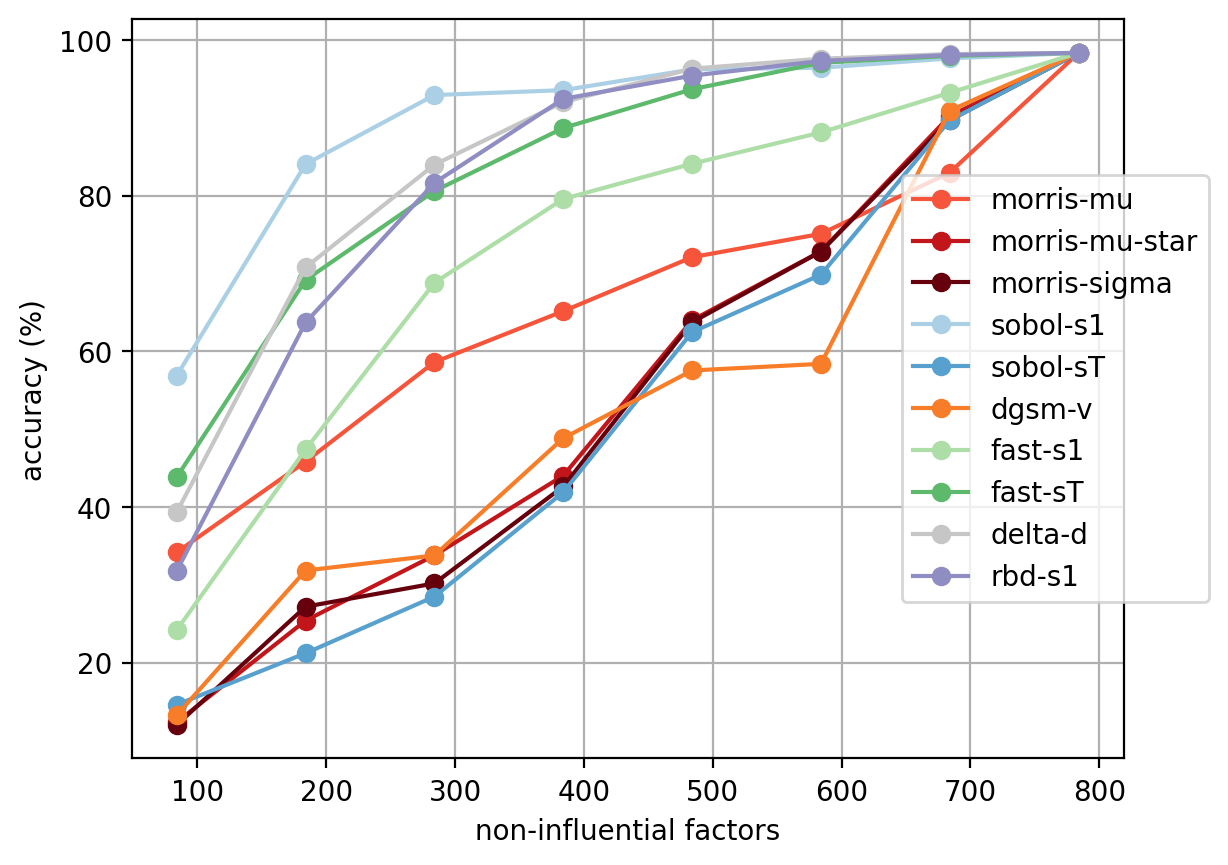}}\label{fig:fig2_b} 
  \caption{Accuracy of MNIST classification. a: b:}
  \label{fig:class_acc}
\end{figure}

\autoref{fig:class_acc} demonstrates the transition between the accuracy of digit classification based on the utilization of the blocks of important features determined by SA methods. A clear incremental trend is observable when more important features are accumulated in ascending manner. The accuracy attained by utilizing the block of the most important features falls below 80\%, whereas it drops below 60\% when employing the blocks of the least important features. In both cases the number of pixels in the first block contains 84 pixels.
 
\begin{figure}[ht]
    \centering
{\includegraphics[width=0.50\textwidth]{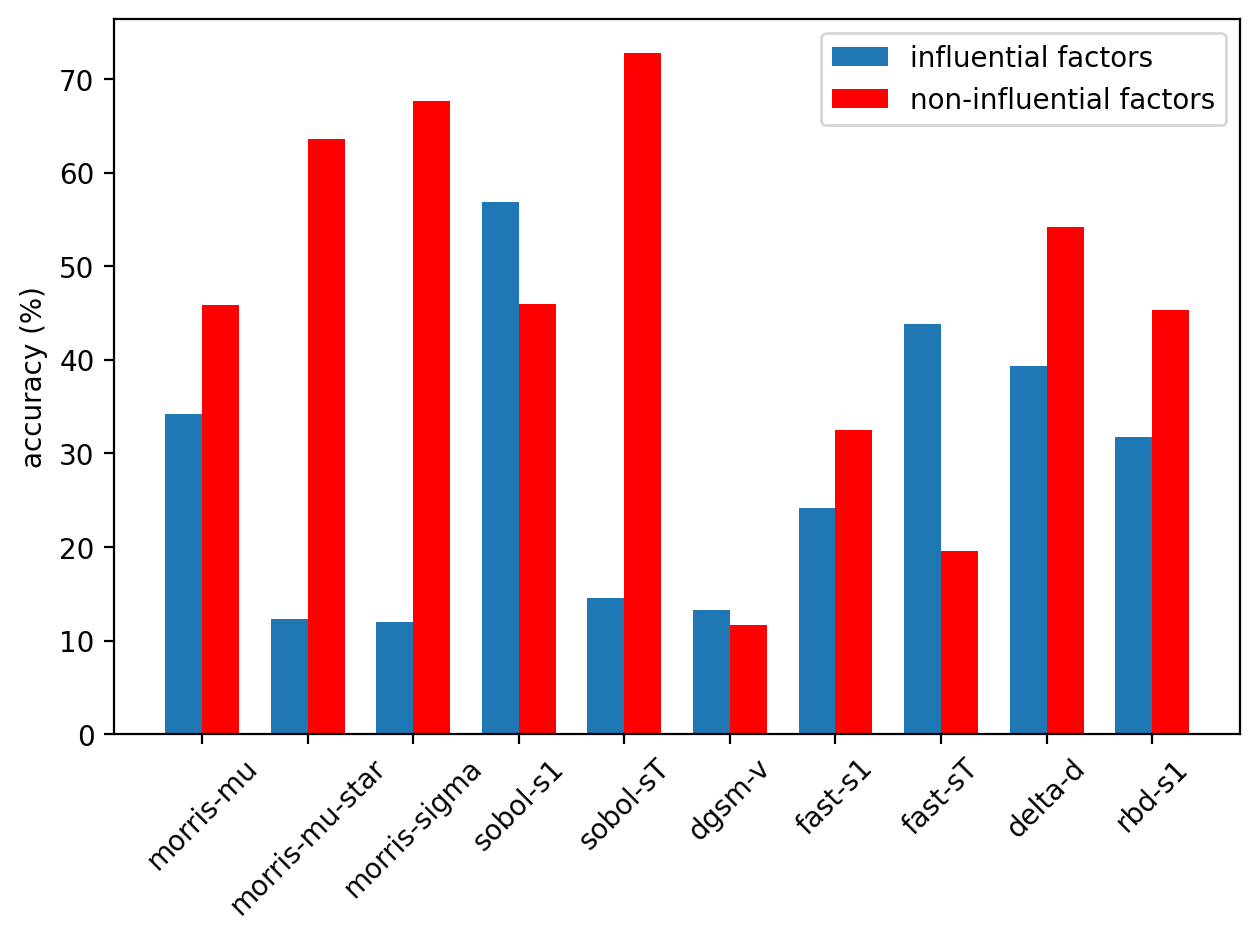}}    \caption{Comparison between the most influential and non-influential factors}
    \label{fig:fig3}
\end{figure}

\begin{figure}[ht]
    \centering
{\includegraphics[width=0.50\textwidth]{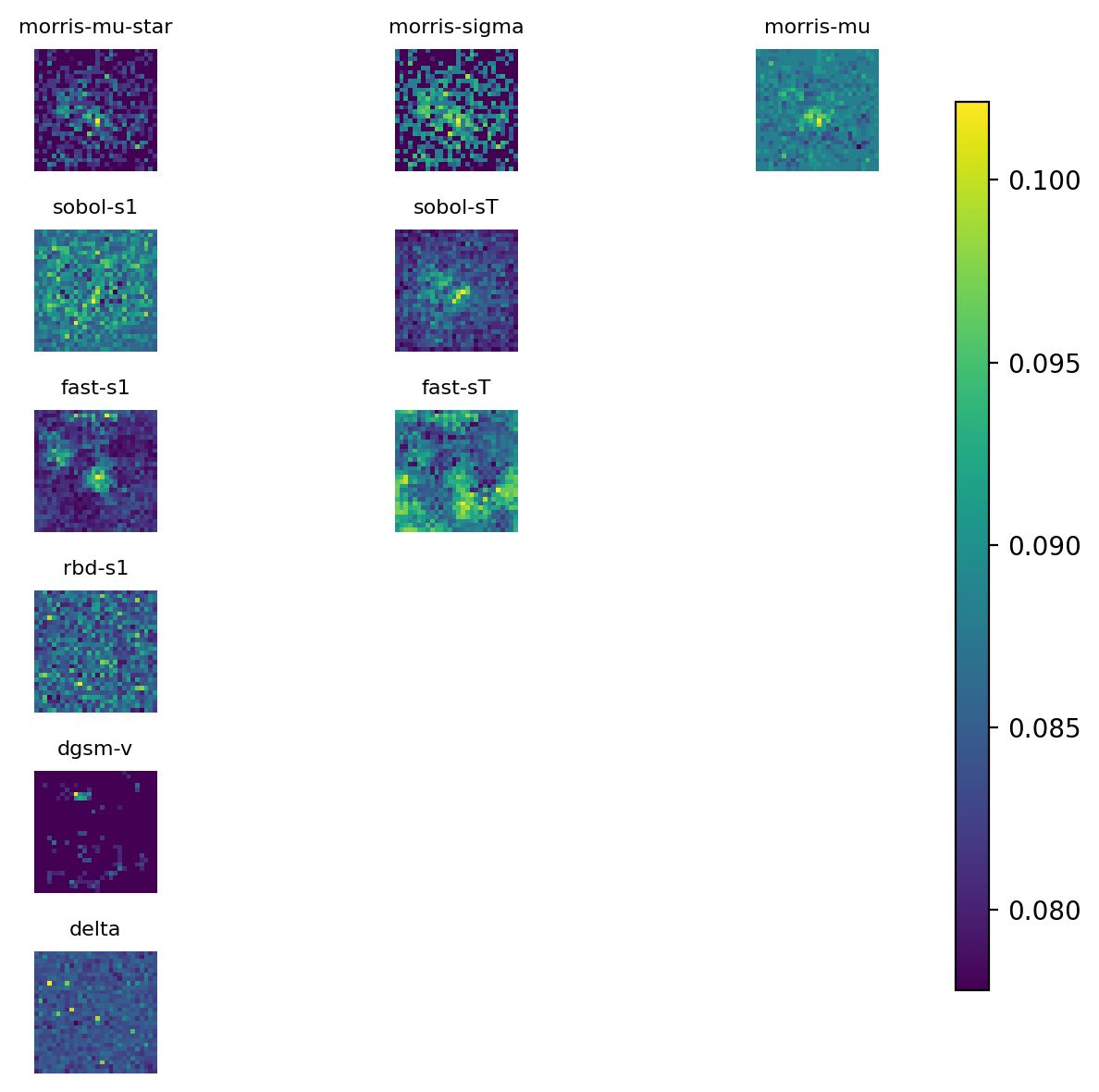}}    \caption{Visualization of the sensitivity indices of each method trained on MNIST classification task}
    \label{fig:fig4}
\end{figure}

Moreover, an increasing pattern can be observed when examining the impact of the unimportant pixels. 
Comparing the results of different methods, it can be concluded that $S_T$ index of Sobol as well as $\sigma$ and $\mu*$ indices of Morris method present superior results. The results affirm the efficacy of the SA indexes in accurately identifying the critical regions that significantly influence correct classification. Moreover, Utilizing the top 684 most important pixels obtained from all the algorithms yields remarkably similar results. Nevertheless, a notable distinction arises in the first 84 significant pixels. Morris method obtains the highest level of accuracy by utilizing the minimum number of most important pixels.  Our simulation results show that $v$ index of DGSM and $S-T$ index of FAST methods are characterized by substantial inconsistency.
A similar conclusion can be derived from \autoref{fig:fig3} which compares the impact of the most influential and non-influential factors. The optimal SA methods are characterized by a substantial accuracy with influential factors and a comparatively lower accuracy with non-influential factors, highlighting a significant disparity in the achieved accuracies between the two factors. It is evident that $\mu*$ and $\sigma$ from Morris and $S_T$ from Sobol demonstrate a significant disparity in accuracy of classification. The obtained sensitivity indices are also visualized in \autoref{fig:fig4} to reveal the pixel locations and their impact on classification results. Since MNIST digits are centered in the images, we expect higher saliency in the central area and lower saliency in the surrounding regions. The results are inline with our previous observation and suggests that the saliency maps returned by Morris, Sobol and FAST specify important pixels more effectively compared to the rest of the discussed methods in this paper. As expected, the different global interpretability methods sometimes yield somewhat different rankings of feature importance when applied to the same machine learning model. Therefore, it is essential to conduct a rigorous analysis for each problem and carefully select the most appropriate global interpretability approach.

\begin{table}[ht]
\centering
\caption{The sampling space size for each of the SA methods}
\begin{tabular}{lll}
\toprule
{} &  \textbf{SA Method} & \textbf{Number of samples} \\
\midrule
  &  Morris &  50 in 4 levels  \\
  &  Sobol &  300   \\
  & FAST & 100 \\
  & RBD & 400 \\
  & Delta & 1000 \\
  & DGSM & 1000 \\
\bottomrule
\end{tabular} 
\end{table}\label{tab:table1}

\bibliographystyle{ieeetr}
\bibliography{ref}

\begin{thebibliography}{10}

\bibitem{aakur2018inherently}
S.~N. Aakur, F.~D. de~Souza, and S.~Sarkar, ``An inherently explainable model for video activity interpretation,'' in {\em Workshops at the Thirty-Second AAAI Conference on Artificial Intelligence}, 2018.

\bibitem{sarvmaili2022modularized}
M.~Sarvmaili, R.~Guidotti, A.~Monreale, A.~Soares, Z.~Sadeghi, F.~Giannotti, D.~Pedreschi, and S.~Matwin, ``A modularized framework for explaining black box classifiers for text data.,'' in {\em Canadian AI}, 2022.

\bibitem{veerappa2022validation}
M.~Veerappa, M.~Anneken, N.~Burkart, and M.~F. Huber, ``Validation of xai explanations for multivariate time series classification in the maritime domain,'' {\em Journal of Computational Science}, vol.~58, p.~101539, 2022.

\bibitem{taylor2024causal}
W.~Taylor-Melanson, Z.~Sadeghi, and S.~Matwin, ``Causal generative explainers using counterfactual inference: A case study on the morpho-mnist dataset,'' {\em arXiv preprint arXiv:2401.11394}, 2024.

\bibitem{van2022comparison}
B.~Van~Stein, E.~Raponi, Z.~Sadeghi, N.~Bouman, R.~C. Van~Ham, and T.~B{\"a}ck, ``A comparison of global sensitivity analysis methods for explainable ai with an application in genomic prediction,'' {\em IEEE Access}, vol.~10, pp.~103364--103381, 2022.

\bibitem{bryan2023explainable}
N.~Bryan-Kinns, C.~Ford, A.~Chamberlain, S.~D. Benford, H.~Kennedy, Z.~Li, W.~Qiong, G.~G. Xia, and J.~Rezwana, ``Explainable ai for the arts: Xaixarts,'' in {\em Proceedings of the 15th Conference on Creativity and Cognition}, pp.~1--7, 2023.

\bibitem{sadeghi2024review}
Z.~Sadeghi, R.~Alizadehsani, M.~A. CIFCI, S.~Kausar, R.~Rehman, P.~Mahanta, P.~K. Bora, A.~Almasri, R.~S. Alkhawaldeh, S.~Hussain, {\em et~al.}, ``A review of explainable artificial intelligence in healthcare,'' {\em Computers and Electrical Engineering}, vol.~118, p.~109370, 2024.

\bibitem{weber2024applications}
P.~Weber, K.~V. Carl, and O.~Hinz, ``Applications of explainable artificial intelligence in finance—a systematic review of finance, information systems, and computer science literature,'' {\em Management Review Quarterly}, vol.~74, no.~2, pp.~867--907, 2024.

\bibitem{fiok2022explainable}
K.~Fiok, F.~V. Farahani, W.~Karwowski, and T.~Ahram, ``Explainable artificial intelligence for education and training,'' {\em The Journal of Defense Modeling and Simulation}, vol.~19, no.~2, pp.~133--144, 2022.

\bibitem{vilone2020comparative}
G.~Vilone, L.~Rizzo, and L.~Longo, ``A comparative analysis of rule-based, model-agnostic methods for explainable artificial intelligence,'' in {\em Proceedings for the 28th AIAI Irish Conference on Artificial Intelligence and Cognitive Science}, Technological University Dublin, 2020.

\bibitem{gan2014comprehensive}
Y.~Gan, Q.~Duan, W.~Gong, C.~Tong, Y.~Sun, W.~Chu, A.~Ye, C.~Miao, and Z.~Di, ``A comprehensive evaluation of various sensitivity analysis methods: A case study with a hydrological model,'' {\em Environmental modelling \& software}, vol.~51, pp.~269--285, 2014.

\bibitem{saltelli2004sensitivity}
A.~Saltelli, S.~Tarantola, F.~Campolongo, M.~Ratto, {\em et~al.}, {\em Sensitivity analysis in practice: a guide to assessing scientific models}, vol.~1.
\newblock Wiley Online Library, 2004.

\bibitem{zhou2018interpreting}
B.~Zhou, D.~Bau, A.~Oliva, and A.~Torralba, ``Interpreting deep visual representations via network dissection,'' {\em IEEE transactions on pattern analysis and machine intelligence}, vol.~41, no.~9, pp.~2131--2145, 2018.

\bibitem{selvaraju2017grad}
R.~R. Selvaraju, M.~Cogswell, A.~Das, R.~Vedantam, D.~Parikh, and D.~Batra, ``Grad-cam: Visual explanations from deep networks via gradient-based localization,'' in {\em Proceedings of the IEEE international conference on computer vision}, pp.~618--626, 2017.

\bibitem{shrikumar2017learning}
A.~Shrikumar, P.~Greenside, and A.~Kundaje, ``Learning important features through propagating activation differences,'' in {\em International conference on machine learning}, pp.~3145--3153, PMLR, 2017.

\bibitem{simonyan2013deep}
K.~Simonyan, A.~Vedaldi, and A.~Zisserman, ``Deep inside convolutional networks: Visualising image classification models and saliency maps,'' {\em arXiv preprint arXiv:1312.6034}, 2013.

\bibitem{sadeghi2019information}
Z.~Sadeghi, ``An information analysis approach into feature understanding of convolutional deep neural networks,'' in {\em Machine Learning, Optimization, and Data Science: 5th International Conference, LOD 2019, Siena, Italy, September 10--13, 2019, Proceedings 5}, pp.~36--44, Springer, 2019.

\bibitem{li2018visualizing}
H.~Li, Z.~Xu, G.~Taylor, C.~Studer, and T.~Goldstein, ``Visualizing the loss landscape of neural nets,'' {\em Advances in neural information processing systems}, vol.~31, 2018.

\bibitem{barrett2023evolutionary}
N.~Barrett, Z.~Sadeghi, and S.~Matwin, ``Evolutionary augmentation policy optimization for self-supervised learning,'' {\em Advances in Artificial Intelligence and Machine Learning Research}, 2023.

\bibitem{pavlick2022semantic}
E.~Pavlick, ``Semantic structure in deep learning,'' {\em Annual Review of Linguistics}, vol.~8, pp.~447--471, 2022.

\bibitem{jingjing2022semantic}
L.~Jingjing, X.~Song, and W.~Lina, ``A semantic interpretation method for deep neural networks based on knowledge graphs,'' in {\em 2022 China Automation Congress (CAC)}, pp.~4665--4668, IEEE, 2022.

\bibitem{sadeghi2016deep}
Z.~Sadeghi, ``Deep learning and developmental learning: emergence of fine-to-coarse conceptual categories at layers of deep belief network,'' {\em Perception}, vol.~45, no.~9, pp.~1036--1045, 2016.

\bibitem{li2023comparison}
D.~Li, P.~Jiang, C.~Hu, and T.~Yan, ``Comparison of local and global sensitivity analysis methods and application to thermal hydraulic phenomena,'' {\em Progress in Nuclear Energy}, vol.~158, p.~104612, 2023.

\bibitem{qin2023comparative}
C.~Qin, Y.~Jin, M.~Tian, P.~Ju, and S.~Zhou, ``Comparative study of global sensitivity analysis and local sensitivity analysis in power system parameter identification,'' {\em Energies}, vol.~16, no.~16, p.~5915, 2023.

\bibitem{reed2022addressing}
P.~Reed, A.~Hadjimichael, K.~Malek, T.~Karimi, C.~Vernon, V.~Srikrishnan, R.~Gupta, D.~Gold, B.~Lee, K.~Keller, {\em et~al.}, ``Addressing uncertainty in multisector dynamics research [book]. zenodo,'' 2022.

\bibitem{chen2021explaining}
H.~Chen, S.~Lundberg, and S.-I. Lee, ``Explaining models by propagating shapley values of local components,'' {\em Explainable AI in Healthcare and Medicine: Building a Culture of Transparency and Accountability}, pp.~261--270, 2021.

\bibitem{ribeiro2016should}
M.~T. Ribeiro, S.~Singh, and C.~Guestrin, ``" why should i trust you?" explaining the predictions of any classifier,'' in {\em Proceedings of the 22nd ACM SIGKDD international conference on knowledge discovery and data mining}, pp.~1135--1144, 2016.

\bibitem{saltelli2002sensitivity}
A.~Saltelli, ``Sensitivity analysis for importance assessment,'' {\em Risk analysis}, vol.~22, no.~3, pp.~579--590, 2002.

\bibitem{molnar2022interpretable}
C.~Molnar, ``Interpretable machine learning, christoph molnar,'' 2022.

\bibitem{molnar2020interpretable}
C.~Molnar, {\em Interpretable machine learning}.
\newblock Lulu. com, 2020.

\bibitem{sobol2001global}
I.~M. Sobol, ``Global sensitivity indices for nonlinear mathematical models and their monte carlo estimates,'' {\em Mathematics and computers in simulation}, vol.~55, no.~1-3, pp.~271--280, 2001.

\bibitem{saltelli2002making}
A.~Saltelli, ``Making best use of model evaluations to compute sensitivity indices,'' {\em Computer physics communications}, vol.~145, no.~2, pp.~280--297, 2002.

\bibitem{saltelli2010variance}
A.~Saltelli, P.~Annoni, I.~Azzini, F.~Campolongo, M.~Ratto, and S.~Tarantola, ``Variance based sensitivity analysis of model output. design and estimator for the total sensitivity index,'' {\em Computer physics communications}, vol.~181, no.~2, pp.~259--270, 2010.

\bibitem{chan1997sensitivity}
K.~Chan, A.~Saltelli, and S.~Tarantola, ``Sensitivity analysis of model output: variance-based methods make the difference,'' in {\em Proceedings of the 29th conference on Winter simulation}, pp.~261--268, 1997.

\bibitem{xu2008general}
C.~Xu and G.~Z. Gertner, ``A general first-order global sensitivity analysis method,'' {\em Reliability Engineering \& System Safety}, vol.~93, no.~7, pp.~1060--1071, 2008.

\bibitem{tarantola2006random}
S.~Tarantola, D.~Gatelli, and T.~A. Mara, ``Random balance designs for the estimation of first order global sensitivity indices,'' {\em Reliability Engineering \& System Safety}, vol.~91, no.~6, pp.~717--727, 2006.

\bibitem{tissot2012bias}
J.-Y. Tissot and C.~Prieur, ``Bias correction for the estimation of sensitivity indices based on random balance designs,'' {\em Reliability Engineering \& System Safety}, vol.~107, pp.~205--213, 2012.

\bibitem{campolongo2007effective}
F.~Campolongo, J.~Cariboni, and A.~Saltelli, ``An effective screening design for sensitivity analysis of large models,'' {\em Environmental modelling \& software}, vol.~22, no.~10, pp.~1509--1518, 2007.

\bibitem{morris1991factorial}
M.~D. Morris, ``Factorial sampling plans for preliminary computational experiments,'' {\em Technometrics}, vol.~33, no.~2, pp.~161--174, 1991.

\bibitem{sobol2010derivative}
I.~M. Sobol and S.~Kucherenko, ``Derivative based global sensitivity measures,'' {\em Procedia-Social and Behavioral Sciences}, vol.~2, no.~6, pp.~7745--7746, 2010.

\bibitem{borgonovo2007new}
E.~Borgonovo, ``A new uncertainty importance measure,'' {\em Reliability Engineering \& System Safety}, vol.~92, no.~6, pp.~771--784, 2007.

\end{thebibliography}

\end{document}